\pdfoutput=1
\documentclass[onecolumn,british,twocolumn]{IEEEtran}
\usepackage[T1]{fontenc}
\usepackage[utf8]{inputenc}
\usepackage{babel}
\usepackage{cprotect}
\usepackage{amsmath}
\usepackage{amsthm}
\usepackage{amssymb}
\usepackage{graphicx}
\usepackage[bookmarks=true,bookmarksnumbered=false,bookmarksopen=false,
 breaklinks=false,pdfborder={0 0 0},pdfborderstyle={},backref=false,colorlinks=false]
 {hyperref}
\hypersetup{pdftitle={Crowded at the Frontier: Model Advantage and Herding among LLM Agents in the Minority Game},
 pdfauthor={Zan-Kai Chong, Alan Yi-Lun Khor, Nur Alysha Ismady},
 hidelinks}

\makeatletter

\providecommand{\tabularnewline}{\\}

\renewenvironment{description}{\list{}{\labelwidth 0pt
 \itemindent -\leftmargin
 }}{\endlist}

\makeatother

\providecommand\theoremname{Theorem}
\theoremstyle{plain}
\newtheorem{thm}{\protect\theoremname}
\providecommand\propositionname{Proposition}
\newtheorem{prop}[thm]{\protect\propositionname}
\providecommand\remarkname{Remark}
\theoremstyle{remark}
\newtheorem{rem}[thm]{\protect\remarkname}
\providecommand\corollaryname{Corollary}
\theoremstyle{plain}
\newtheorem{cor}[thm]{\protect\corollaryname}
\usepackage{listings}

\begin{document}
\title{Crowded at the Frontier: Model Advantage and Herding among LLM Agents
in the Minority Game}
\author{Zan-Kai Chong, Alan Yi-Lun Khor and Nur Alysha Ismady\\
{\normalsize\itshape Independent Researchers}\\
{\normalsize\itshape Kuala Lumpur, Malaysia}\\
{\normalsize zkchong@gmail.com, alankhor96@gmail.com, alyshaismady01@gmail.com}}
\maketitle
\begin{abstract}
As LLM agents are deployed to compete for shared, finite resources,
a natural question arises. What is the value of having the best model
when many competitors have it too? We study this in the Minority Game,
where agents win by choosing the less crowded side and predictability
is penalised. We build mixed populations of \emph{frontier} and \emph{secondary}
LLM agents with access to $m$ bits of public history, and sweep the
frontier share $f=k/N$. We hypothesise that the frontier advantage
$\Delta(f)$ is positive when frontier agents are rare and falls beyond
a critical share $f^{*}$. Two analytical benchmarks, in which frontier
agents either herd or share the minority side, give the same closed-form
threshold $f^{*}=2\epsilon/(2\epsilon+2q-1)$ in terms of the secondary
crowd's exploitability $\epsilon$ and the frontier model's reliability
$q$, but they disagree beyond it. In that range, herding predicts
a reversal, with frontier agents losing to secondary agents, and sharing
predicts a positive edge of $2\epsilon/f$. A private-randomness ablation
tests whether lowering the correlation among frontier agents restores
the advantage. The benchmarks show that randomisation at a rate tuned
to the share converts herding into sharing. A pilot with 11 agents,
Gemini 3.1 Pro against Gemini 3.1 Flash-Lite, finds a small and noisy
advantage for a single frontier agent, none at shares of 0.18 and
0.27, and a reversal at 0.55, as herding predicts. The share at which
the advantage vanishes, about 0.13, is close to the 0.14 predicted
from the estimated parameters, although that prediction has a 95\%
range of 0.07 to 0.25 at this size. The correlation among frontier
agents is close to the benchmark's $(2q-1)^{2}$. A control in which
the shared rule ignores the crowd reproduces the reversal and this
correlation, and the evidence that the frontier model reads the crowd
is weaker. Private randomness lowers the correlation at two of those
three shares but does not restore the advantage. A fixed rate of randomisation
would have this effect in the benchmark, but the call logs show that
few choices follow the number. If this pattern holds for other models
and larger populations, once frontier models are widely adopted, being
different matters more than being best.
\end{abstract}

\section{Introduction}\label{sec:intro}

Frontier language models are produced by a small number of large technology
companies and are increasingly used as autonomous decision-makers
in markets, scheduling, routing, and resource allocation. When one
firm adopts a frontier model and its competitors do not, it may gain
a clear edge. But adoption spreads. If most competitors run the same
or similar frontier models, their decisions may converge, and the
advantage of the better model may vanish, or even reverse.

The Minority Game is the canonical model for this kind of anti-coordination.
Arthur \cite{arthur1994} introduced the El Farol Bar problem, and
Challet and Zhang \cite{challet1997} formalised it as the Minority
Game. Agents repeatedly choose one of two actions, and those on the
minority side win. Success requires being different from the crowd,
and any shared reasoning pattern is self-defeating. Classical theory
shows that collective efficiency depends on strategic diversity and
on the ratio of information to population size, and it measures efficiency
by the volatility. Crowd--anti-crowd theory \cite{johnson1999} attributes
the volatility to groups of agents that act alike, and Challet, Marsili
and Ottino \cite{challet2004} show that when agents have too little
information, crowd effects can make coordination worse than that of
random agents.

The idea that an edge erodes once it is widely shared is not new.
We call it dilution when the edge falls to zero, and reversal when
it turns into a loss. Producer--speculator models \cite{challet2000}
show that speculators' gains shrink once their number grows large,
and mixed-memory populations \cite{johnson1999mixed} show that the
advantage of longer memory depends on its prevalence. In economics,
Grossman and Stiglitz \cite{grossman1980} show that the return to
costly information falls as more traders acquire it. Kleinberg and
Raghavan \cite{kleinberg2021} study algorithmic monoculture, in which
many decision-makers rely on the same algorithm. They show that collective
outcomes can worsen even when that algorithm is more accurate for
each decision-maker alone.

LLM agents have been studied in repeated matrix games \cite{akata2025}
and in multi-agent gaming environments \cite{huang2025}, which include
an El Farol game played by ten copies of one model. Two studies of
the El Farol problem are closest in setting. Mori and Johnson \cite{mori2026}
find that LLM populations coordinate worse than calibrated random
agents and that larger models overload more often, though model size,
prompts, and exploration are confounded and populations contain only
three or four agents. Takata et al. \cite{takata2025} study 20 GPT-4o
agents in a spatial El Farol setting and report human-like, socially
motivated satisficing. Neither study mixes model tiers within a population
or measures \emph{relative} advantage.

Three recent studies are closest in mechanism. Ballestero et al. \cite{ballestero2026}
pair each of 16 LLMs with a copy of itself in two-player games, and
the copies still agree 27\% of the time when divergence is rewarded,
against 4\% for humans. Sinishaw et al. \cite{sinishaw2026} tell
agents that they face copies of themselves in one-shot games. The
copies do well in common-interest games, but several open-weight models
fall below the Nash benchmark in anti-coordination games. Ross et
al. \cite{ross2026} show that the correlation between the actions
of different LLMs rises with capability, and they sweep the share
of LLM traders against noise traders to measure market risk.

The clearest evidence that copies of one model fail to diverge comes
from one-shot games. A repeated game gives agents more to work with,
because they see a shared history of outcomes. They could use it to
spread across the two sides and keep a shrinking edge, or they could
move together and lose. None of the studies above builds a population
from two model tiers, varies the share of the stronger model, or compares
their payoffs. It is therefore open which outcome holds for LLM agents,
at what share the edge vanishes, and whether correlated actions rather
than a lack of capability explain the decline. We ask three questions:
\begin{enumerate}
\item \textbf{Dilution.} How does the frontier model's advantage over a
secondary model change as the share of frontier agents rises? Is there
a critical share $f^{*}$?
\item \textbf{Reversal.} If so, beyond $f^{*}$, do frontier agents do worse
than secondary agents because they herd, or do they share the minority
side and keep a shrinking edge?
\item \textbf{Mechanism.} Is the decline caused by frontier agents acting
alike, such that breaking that symmetry restores the advantage without
changing the model?
\end{enumerate}
This paper makes three contributions. First, to our knowledge, it
gives the first measurement of a frontier model's relative advantage
$\Delta(f)$ as a function of its adoption share in a mixed LLM population.
It also derives two analytical benchmarks that share a closed-form
threshold $f^{*}$ and make opposite predictions beyond it, so that
the experiment discriminates between herding and sharing. Second,
it gives evidence that the decline is consistent with correlated actions
among identical frontier agents, measured through the correlation
$\rho_{\text{FF}}$, in a repeated population game where the cost
of correlation depends on how many agents share it. This extends the
one-shot, small-group findings of Ballestero et al. \cite{ballestero2026}
and Sinishaw et al. \cite{sinishaw2026} to a repeated game in a mixed
population, and it tests a private-randomness intervention designed
to lower the correlation without changing the model. Third, it shows
that private randomness converts herding into sharing only at a rate
that rises with the frontier share. In the pilot, private randomness
did not restore the advantage, and the call logs show that few choices
follow the number. The measurements come from a pilot with 11 agents
and one model pair, and Section \ref{sec:results} states which hypotheses
it supports and which remain open.

\section{Game and Agents}\label{sec:model}

\subsection{Game and payoffs}\label{sec:game}

At each round $t$, each agent $i\in\{1,\dots,N\}$ chooses $a_{i}(t)\in\{-1,+1\}$,
with $N$ odd. Let $A(t)=\sum_{i}a_{i}(t)$ denote the net vote. Its
sign gives the majority side. The minority side $-\mathrm{sgn}\,A(t)$
wins. Winning agents receive $+1$ and losing agents $-1$. The public
history $\mu(t)$ is the string of the last $m$ winning sides, and
there are therefore $2^{m}$ possible histories. All agents move simultaneously,
as in the classic Minority Game, and no agent sees another agent's
current choice.

Let $\bar{w}_{\text{F}}$ and $\bar{w}_{\text{S}}$ denote the mean
payoff per round of the frontier and secondary groups. The frontier
advantage is $\Delta(f)=\bar{w}_{\text{F}}(f)-\bar{w}_{\text{S}}(f)$,
and it is defined for $0<f<1$ because both groups must be present.

\subsection{Agents and prompt}\label{sec:populations}

The population has $k$ \emph{frontier agents}, which run a frontier
LLM, and $N-k$ \emph{secondary agents}, which run a smaller LLM from
the same family. The frontier share is $f=k/N$. All agents receive
the same neutral, framing-free prompt. Agents are not told which model
other agents use. Each agent writes a short rationale before choosing,
which other agents do not see.

Listing \ref{lst:prompt} reproduces the prompt. Square brackets mark
the values filled in each round. Side A maps to $+1$ and side B to
$-1$. The last line containing the word CHOICE is parsed. A response
without a parseable choice counts as a parse failure, and the agent
plays a fair coin that round. In the private randomness condition
the line of Listing \ref{lst:random} is inserted after the score
line.

\bgroup\inputencoding{latin9}
\begin{lstlisting}[basicstyle={\footnotesize\ttfamily},breaklines=true,frame=single,captionpos=b,float=tb,aboveskip=6pt,belowskip=6pt,abovecaptionskip=4pt,belowcaptionskip=2pt,caption={Prompt given to every agent in every round. Bracketed fields are filled in per agent and per round.},label={lst:prompt}]
You are one of [N] players in a repeated game. Each round, every player privately chooses side A or side B at the same time. After all choices are made, the side chosen by fewer players wins. Each player on the winning side gets +1, each player on the losing side gets -1. The same [N] players play every round.

Information available to you:
- The winning sides of the last [m] rounds, oldest first: [history]
- Your own choices and payoffs in the last [m] rounds, oldest first: [record]
- Your total score so far: [score]

Think briefly about what to do (at most three sentences), then answer in exactly this format on the last line:
CHOICE: A
or
CHOICE: B
\end{lstlisting}
\leavevmode\egroup{}

\bgroup\inputencoding{latin9}
\begin{lstlisting}[basicstyle={\footnotesize\ttfamily},breaklines=true,frame=single,captionpos=b,float=tb,aboveskip=6pt,belowskip=6pt,abovecaptionskip=4pt,belowcaptionskip=2pt,caption={Line added in the private randomness condition. The random number u is printed with four decimals.},label={lst:random}]
- A private random number drawn uniformly between 0 and 1 for you this round: [u]. No other player sees it. You may use it in any way you like, or ignore it.
\end{lstlisting}
\leavevmode\egroup{}

\section{Analytical Benchmarks}\label{sec:benchmarks}

Before running agents, we derive two stylised benchmarks for $\Delta(f)$.
Both share the same population setup, but they differ in how frontier
agents treat one another. In the \emph{herding} benchmark, every frontier
agent reacts to the same signal and ignores the fact that the others
do too. In the \emph{sharing} benchmark, frontier agents anti-coordinate
among themselves as well as the game allows. The two benchmarks agree
below a threshold $f^{*}$ and disagree beyond it. LLM agents are
expected to fall between them, and the experiment measures where they
fall.

\subsection{Setup}\label{sec:bench-setup}

Fix a history $\mu$ and let $b=b(\mu)\in\{-1,+1\}$ denote the side
towards which a predictable crowd leans. In the experiment, $b(\mu)$
is the sign of the secondary crowd's mean action given $\mu$. A secondary
agent plays $b$ with probability $\tfrac{1}{2}+\epsilon$, where
$\epsilon\in(0,\tfrac{1}{2})$ measures how exploitable the crowd
is. A frontier agent identifies the lean and plays $-b$ with probability
$q\in(\tfrac{1}{2},1]$, and plays $b$ otherwise, where $q$ is the
reliability of the frontier model, our measure of its capability.
Conditional on $\mu$, all actions are independent.

These two rules fix the mean action of each group. A secondary action
equals $b$ with probability $\tfrac{1}{2}+\epsilon$ and $-b$ otherwise,
and its mean is therefore $(\tfrac{1}{2}+\epsilon)b-(\tfrac{1}{2}-\epsilon)b=2\epsilon b$.
Likewise a frontier action equals $-b$ with probability $q$ and
$b$ otherwise, and its mean is $-qb+(1-q)b=-(2q-1)b$. Writing the
condition on the history explicitly,
\begin{equation}
\mathbb{E}[a_{\text{S}}\mid\mu]=2\epsilon b,\qquad\mathbb{E}[a_{\text{F}}\mid\mu]=-(2q-1)b.\label{eq:cond-means}
\end{equation}
By analogy with \cite{challet2000stat}, we call the size of a group's
conditional mean its predictability,
\begin{equation}
h_{g}(\mu)=\bigl|\mathbb{E}[a_{g}\mid\mu]\bigr|.\label{eq:predictability}
\end{equation}
By (\ref{eq:cond-means}), $h_{\text{S}}(\mu)=2\epsilon$ for the
crowd and $h_{\text{F}}(\mu)=2q-1$ for the frontier bloc.

With both conditional means in hand, we can form the expected net
vote. Summing (\ref{eq:cond-means}) over the $(1-f)N$ secondary
agents and the $k=fN$ frontier agents gives $\mathbb{E}[A]=(1-f)N\cdot2\epsilon b-fN(2q-1)b$.
Factoring out $Nb$, we write $\mathbb{E}[A]=NbD(f)$, where

\begin{equation}
D(f)=2\epsilon(1-f)-(2q-1)f.\label{eq:drift}
\end{equation}

The drift $D$ is the expected net vote per agent, measured along
the lean $b$. Its first term is the crowd's pull towards $b$, weak
per agent but carried by a large share, and its second term is the
frontier bloc's pull away from $b$, strong per agent but carried
by a small share. Therefore $D$ is positive when the crowd outweighs
the frontier bloc and negative when the bloc outweighs the crowd.
As $f$ rises, $D$ falls linearly from $2\epsilon$ at $f=0$ to
$-(2q-1)$ at $f=1$. The two pulls therefore balance at a single
share,

\begin{equation}
f^{*}=\frac{2\epsilon}{2\epsilon+2q-1}.\label{eq:fstar}
\end{equation}

Because $D>0$ for $f<f^{*}$, the minority side is $-b$ and frontier
agents who identify the lean win.

\subsection{Herding benchmark}\label{sec:bench-herding}

The first benchmark supposes that every frontier agent keeps the rule
of the setup, however many of them there are. We call it the herding
benchmark, because agents that follow one rule move together.
\begin{prop}[Herding]
\label{prop:herding}Suppose frontier agents play as in the setup
for every $f$. Then as $N\to\infty$ with $f$ fixed and $f\neq f^{*}$,
the expected payoffs per round of a frontier and a secondary agent
are
\begin{equation}
w_{\text{F}}(f)=(2q-1)\,\mathrm{sgn}\,D(f),\label{eq:herding-wF}
\end{equation}
\begin{equation}
w_{\text{S}}(f)=-2\epsilon\,\mathrm{sgn}\,D(f),\label{eq:herding-wS}
\end{equation}
and therefore
\begin{equation}
\Delta(f)=(2q-1+2\epsilon)\,\mathrm{sgn}\,D(f).\label{eq:herding}
\end{equation}
The advantage $\Delta$ is therefore positive for $f<f^{*}$ and negative
for $f>f^{*}$. It does not fade but switches from its full value
to an equal loss at $f^{*}$, and dilution and reversal occur at the
same share.
\end{prop}
\begin{IEEEproof}
By the weak law of large numbers, $A/N$ converges in probability
to $bD(f)$. Because $D(f)\neq0$, $\mathrm{sgn}\,A$ converges in
probability to $b\,\mathrm{sgn}\,D(f)$. An agent's payoff is $-a_{i}\,\mathrm{sgn}\,A$
and is bounded by 1. Its expectation therefore converges to $-\mathbb{E}[a_{i}]\,b\,\mathrm{sgn}\,D(f)$.
Substituting the conditional means (\ref{eq:cond-means}) gives (\ref{eq:herding-wF}),
(\ref{eq:herding-wS}) and (\ref{eq:herding}). The sign claims follow
from the definition of $f^{*}$ in (\ref{eq:fstar}).
\end{IEEEproof}
\begin{rem}[Finite populations]
\label{rem:finite-n}For finite $N$, $A$ is a sum of independent
$\pm1$ variables with mean $NbD(f)$ and variance $Nv(f)$, where
$v(f)=(1-f)(1-4\epsilon^{2})+f\bigl(1-(2q-1)^{2}\bigr)$. The normal
approximation of $A$ replaces $\mathrm{sgn}\,D(f)$ in (\ref{eq:herding})
by $\mathbb{E}[b\,\mathrm{sgn}\,A]=2\Phi\bigl(\sqrt{N}\,D(f)/\sqrt{v(f)}\bigr)-1$
and gives
\begin{equation}
\Delta_{N}(f)\approx(2q-1+2\epsilon)\Bigl[2\Phi\Bigl(\tfrac{\sqrt{N}\,D(f)}{\sqrt{v(f)}}\Bigr)-1\Bigr],\label{eq:herding-finite}
\end{equation}
a sigmoid in $f$ centred at $f^{*}$ with width of order $1/\sqrt{N}$
that returns to (\ref{eq:herding}) as $N\to\infty$. For the illustrative
values of Fig. \ref{fig:benchmarks}, exact enumeration of the finite
game agrees with (\ref{eq:herding-finite}) to within about 0.1 for
$N=11$ and within 0.03 for $N=101$. It crosses zero later than (\ref{eq:herding-finite})
does, at $f=0.30$ for $N=11$, 0.27 for $N=31$ and 0.26 for $N=101$,
against $f^{*}=0.25$. The gap is about $0.5/N$ and vanishes as $N\to\infty$.
\end{rem}

\subsection{Sharing benchmark}\label{sec:bench-sharing}

The herding benchmark assumes frontier agents never respond to one
another. The opposite extreme lets them anti-coordinate perfectly.
We call it the sharing benchmark, because frontier agents share the
minority side among themselves. Frontier agents that identify the
lean, a fraction $q$ of them in expectation, may place themselves
on either side in any way, while the remaining fraction $1-q$ still
plays $b$. Secondary agents are unchanged.

This benchmark is a ceiling rather than a prediction. Beyond $f^{*}$,
reaching it requires that some frontier agents sit on the losing side
in a given round. No such agent can gain by moving, because one more
agent on $-b$ tips it into the majority, and the allocation is therefore
stable once reached. But nothing in the one-shot game says who takes
the losing side. In repeated play, rotation or the private randomisation
of Corollary \ref{cor:randomisation} breaks that symmetry, and every
frontier agent then earns the same payoff in expectation.
\begin{prop}[Sharing]
\label{prop:sharing}Under these assumptions, as $N\to\infty$ the
largest expected frontier payoff is
\begin{equation}
w_{\text{F}}(f)=\begin{cases}
2q-1, & f\le f^{*},\\[2pt]
\dfrac{2\epsilon(1-f)}{f}, & f>f^{*},
\end{cases}\label{eq:sharing-w}
\end{equation}
with $w_{\text{S}}=-2\epsilon$ throughout. This gives
\begin{equation}
\Delta(f)=\begin{cases}
2q-1+2\epsilon, & f\le f^{*},\\[2pt]
\dfrac{2\epsilon}{f}, & f>f^{*}.
\end{cases}\label{eq:sharing}
\end{equation}
$\Delta$ is continuous and non-increasing, and it stays positive
for every $f<1$.
\end{prop}
\begin{IEEEproof}
As $N\to\infty$ the fraction of secondary agents on $-b$ converges
to $(1-f)(\tfrac{1}{2}-\epsilon)$ by the weak law of large numbers,
and we count agents as fractions of $N$. The side $-b$ remains the
minority only while it holds at most $\tfrac{1}{2}$, and the room
left for identifying frontier agents is therefore $R(f)=\tfrac{1}{2}-(1-f)(\tfrac{1}{2}-\epsilon)=\tfrac{1}{2}f+\epsilon(1-f)$.
These agents number $qf$. The condition $qf\le R(f)$ rearranges
to $(2q-1)f\le2\epsilon(1-f)$, which is $D(f)\ge0$ and therefore
$f\le f^{*}$. In that case all of them win and $w_{\text{F}}=q-(1-q)=2q-1$.
If $qf>R(f)$, at most $R(f)$ frontier agents can sit on $-b$ while
it stays the minority. The best allocation fills the room $R(f)$
exactly, because each frontier agent placed there wins. The other
$f-R(f)$ frontier agents lose. They include those that do not identify
the lean. The frontier payoff per agent is then $\bigl[R-(f-R)\bigr]/f=(2R-f)/f=2\epsilon(1-f)/f$.

Letting $-b$ become the majority instead is worse. Because $R(f)$
frontier agents already fill $-b$ to one half, tipping it into the
majority puts more than $R(f)$ of them on the losing side. The total
frontier payoff is then at most $f-2R(f)=-2\epsilon(1-f)<0$, whereas
filling the minority side yields $+2\epsilon(1-f)$. Secondary agents
sit on $-b$ with probability $\tfrac{1}{2}-\epsilon$ and $-b$ is
the minority, giving $w_{\text{S}}=-2\epsilon$. Continuity at $f^{*}$
follows from (\ref{eq:fstar}), and $2\epsilon/f$ is decreasing and
positive.
\end{IEEEproof}
\begin{rem}[Capability no longer matters]
\label{rem:ceiling}Beyond $f^{*}$, $w_{\text{F}}$ does not depend
on $q$. Once the bloc is large enough to fill the minority side,
capability no longer sets the ceiling. Only the crowd's exploitability
$\epsilon$ and the bloc's size do.
\end{rem}

\subsection{Private randomness}\label{sec:bench-randomness}

Sharing requires frontier agents to coordinate, which is hard when
they move at the same time and do not communicate. Each agent can,
however, loosen the bloc on its own by sometimes replacing its choice
with a private coin flip.
\begin{cor}[Randomisation]
\label{cor:randomisation}In the herding benchmark, let each frontier
agent replace its action by a fair coin with probability $r$. Then
$2q-1$ becomes $(1-r)(2q-1)$ and
\begin{equation}
f^{*}(r)=\frac{2\epsilon}{2\epsilon+(1-r)(2q-1)},\label{eq:fstar-r}
\end{equation}
which increases with $r$. For $f>f^{*}$, the frontier payoff is
$(1-r)(2q-1)$ whenever $r$ exceeds
\begin{equation}
r^{*}(f)=1-\frac{2\epsilon(1-f)}{(2q-1)f},\label{eq:rstar}
\end{equation}
and its supremum over such $r$ equals the sharing benchmark $2\epsilon(1-f)/f$.
\end{cor}
\begin{IEEEproof}
Replacing a fraction $r$ of actions by fair coins scales the conditional
mean of a frontier action in (\ref{eq:cond-means}) by $1-r$ and
leaves the secondary mean unchanged. Substituting into (\ref{eq:fstar})
gives (\ref{eq:fstar-r}). For $r>r^{*}(f)$ the drift is positive,
and Proposition \ref{prop:herding} therefore applies with the scaled
mean. Moreover, $(1-r)(2q-1)\to2\epsilon(1-f)/f$ as $r$ decreases
to $r^{*}(f)$.
\end{IEEEproof}
\begin{rem}[Too many frontier agents]
\label{rem:too-many}Corollary \ref{cor:randomisation} tunes $r$
to $f$. Suppose instead that every frontier agent randomises at one
fixed rate $r$. Proposition \ref{prop:herding} then applies with
the scaled mean for every $f$. The advantage $\Delta$ equals $(1-r)(2q-1)+2\epsilon$
below $f^{*}(r)$ and the negative of that value above it. A fixed
rate therefore moves the reversal from $f^{*}$ to $f^{*}(r)$ and
shrinks it, but it does not remove it. Only a rate that rises with
the share, $r>r^{*}(f)$, keeps the edge positive at every $f$. That
rate approaches 1 as $f\to1$, and the ceiling $2\epsilon(1-f)/f$
approaches 0. In a population that is almost all frontier agents,
the model avoids a loss only by being ignored, and the payoff it retains
is almost nothing.
\end{rem}
\begin{rem}[Randomising crowd]
\label{rem:random-crowd}If secondary agents also flip the coin,
at a rate $r_{\text{S}}$, then $\epsilon$ scales to $(1-r_{\text{S}})\epsilon$
in the same way, and the threshold rises only if $r>r_{\text{S}}$.
\end{rem}
\begin{figure}
\begin{centering}
\includegraphics[width=1\columnwidth]{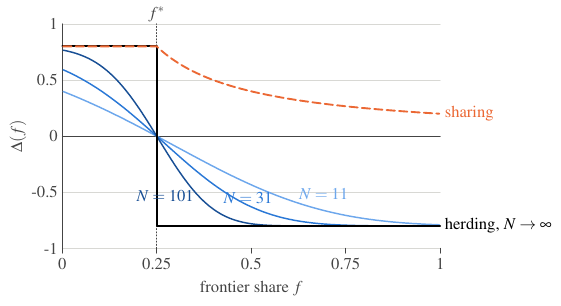}
\par\end{centering}
\caption{Predicted frontier advantage $\Delta(f)$ for illustrative values
$\epsilon=0.1$ and $q=0.8$, which give $f^{*}=0.25$. Herding (Proposition
\ref{prop:herding}): the $N\to\infty$ step and its finite-$N$ form
(\ref{eq:herding-finite}) for $N=11,31,101$. Sharing (Proposition
\ref{prop:sharing}): the upper bound $2\epsilon/f$. The benchmarks
coincide below $f^{*}$ and disagree in sign beyond it.}\label{fig:benchmarks}
\end{figure}

\subsection{Predictions}\label{sec:bench-predictions}

This subsection draws out what the benchmarks imply for the experiment,
in the form of three tests. Fig. \ref{fig:benchmarks} plots the herding
and sharing benchmarks for illustrative values, $\epsilon=0.1$ and
$q=0.8$, and the first test can be read off it.

\subsubsection{Sign of the advantage}\label{sec:test-sign}

For $f<f^{*}$ the benchmarks agree. Both give the same $f^{*}$ and
the same constant advantage $\Delta(f)=2q-1+2\epsilon$, which we
call the plateau. Dilution alone therefore does not separate them.
For $f>f^{*}$ they part. Herding predicts $\Delta=-(2q-1+2\epsilon)$
and sharing predicts $\Delta=2\epsilon/f>0$, and the sign of $\Delta$
beyond the threshold is the first test.

Proposition \ref{prop:sharing} holds as $N\to\infty$. At finite
$N$ we compute its counterpart by enumeration. The frontier agents
that identify the lean learn their own number. They then choose how
many of them take side $-b$ so as to maximise the bloc's expected
payoff, without seeing the crowd's choices. With a single frontier
agent there is nothing to coordinate, and the two benchmarks coincide.
With two or more they separate, first in size and then, at larger
$f$, in sign.

Population size also sets how sharp that first test is. At $N=11$
the finite-$N$ form (\ref{eq:herding-finite}) is a wide sigmoid,
and herding predicts only a small edge at small $k$. At $N=101$
the step is nearly sharp. The crossing point moves less than the shape
does. Approximation (\ref{eq:herding-finite}) crosses zero at $f^{*}$
for every $N$, and the exact game crosses above it by a gap that
is inversely proportional to $N$ (Remark \ref{rem:finite-n}). The
threshold can therefore be estimated at small $N$ if it is compared
with the exact crossing, while the shape of the curve requires large
$N$.

\subsubsection{Predicted threshold and correlations}\label{sec:test-threshold}

The threshold estimate can be checked against a prediction that has
no free parameters. The conditional means (\ref{eq:cond-means}) give
$\epsilon$ from the secondary-only population and $q$ from a single
frontier agent, and (\ref{eq:fstar}) then gives $f^{*}$ in advance.
Agreement between the measured and the predicted $f^{*}$ is the second
test.

The same two parameters also fix the pairwise correlations. Averaging
over histories with $b$ balanced, every action has mean zero and
unit variance, and actions are independent given $b$. Each pairwise
correlation is therefore the product of the two conditional means
in (\ref{eq:cond-means}), giving $\rho_{\text{FF}}=(2q-1)^{2}$,
$\rho_{\text{SS}}=4\epsilon^{2}$ and $\rho_{\text{FS}}=-2\epsilon(2q-1)$.
In terms of the predictability (\ref{eq:predictability}), each group's
correlation is its own predictability squared, $\rho_{\text{FF}}=h^{2}_{\text{F}}$
and $\rho_{\text{SS}}=h^{2}_{\text{S}}$. The cross-correlation $\rho_{\text{FS}}$
is the only one of the three that depends on the direction of the
frontier lean. It therefore tests whether the frontier rule reads
the crowd, because the unaligned rule of Remark \ref{rem:unaligned}
gives zero.
\begin{rem}[Unaligned rule]
\label{rem:unaligned}Proposition \ref{prop:herding} assumes that
the frontier rule opposes the crowd's lean in every history. Suppose
instead that frontier agents share a rule $c(\mu)\in\{-1,+1\}$, play
it with probability $q$, and that $c(\mu)=-b(\mu)$ in only half
the histories. In the other half the bloc and the crowd pull the same
way, the minority side is $-b$, and a frontier agent earns $-(2q-1)$.
Averaging over the two halves as $N\to\infty$ gives $\Delta=2\epsilon$
for $f<f^{*}$ and $\Delta=-(2q-1)$ for $f>f^{*}$. The predictability
$h_{\text{F}}=2q-1$ and the correlation $\rho_{\text{FF}}=(2q-1)^{2}$
are unchanged, but $\rho_{\text{FS}}=0$. The reversal and the frontier
correlation therefore need only a shared rule. The plateau below $f^{*}$
and the sign of $\rho_{\text{FS}}$ are what show that the rule reads
the crowd.
\end{rem}

\subsubsection{Volatility}\label{sec:test-volatility}

The volatility gives a third test, although not an independent one.
Let $\sigma^{2}$ be the variance of the net vote $A$ across rounds.
We report $\sigma^{2}/N$, which equals 1 when every agent flips a
fair coin. Because the net vote is a sum of actions, its variance
follows from the pairwise correlations. When every action has mean
zero,
\begin{equation}
\begin{split}\frac{\sigma^{2}}{N}=1+\frac{1}{N}\bigl[ & k(k-1)\rho_{\text{FF}}+2k(N-k)\rho_{\text{FS}}\\
 & +(N-k)(N-k-1)\rho_{\text{SS}}\bigr].
\end{split}
\label{eq:vol-identity}
\end{equation}
The volatility therefore restates the three correlations at the level
of the population. By Remark \ref{rem:finite-n}, under herding $A$
has mean $NbD(f)$ and variance $Nv(f)$ for a fixed lean, and averaging
over $b$ gives $\sigma^{2}/N=v(f)+N\,D(f)^{2}$. Because $D(f^{*})=0$,
the second term vanishes at $f^{*}$, and the volatility falls to
a minimum near $f^{*}$ and rises on both sides. Under sharing, the
bloc holds the net vote near zero beyond $f^{*}$, and the volatility
stays at most 1. The two benchmarks therefore share the fall to $f^{*}$,
and a rise in $\sigma^{2}/N$ beyond it points to herding.

\section{Experimental Design}\label{sec:design}

\subsection{Hypotheses}\label{sec:hypotheses}

The hypotheses below turn the tests of Section \ref{sec:bench-predictions}
into claims about LLM agents, and Section \ref{sec:results} reports
a verdict on each. H1 and H2 answer the first two questions of Section
\ref{sec:intro}, and H3 and H4 both answer the third.
\begin{description}
\item [{H1~(Dilution).}] $\Delta(f)$ is positive for small $f$ and falls
beyond a critical share $f^{*}<1$, as both benchmarks predict.
\item [{H2~(Reversal,~first~test).}] Beyond $f^{*}$, $\Delta(f)$ is
negative, as in the herding benchmark of Proposition \ref{prop:herding},
rather than positive, as in the sharing benchmark of Proposition \ref{prop:sharing}.
\item [{H3~(Mechanism,~second~and~third~tests).}] Frontier agents
share a lean that opposes the crowd. The correlations $\rho_{\text{FF}}$
and $\rho_{\text{FS}}$ are close to $(2\hat{q}-1)^{2}$ and $-2\hat{\epsilon}(2\hat{q}-1)$,
the measured $\hat{f}^{*}$ agrees with (\ref{eq:fstar}) evaluated
at $\hat{\epsilon}$ and $\hat{q}$, and $\sigma^{2}/N$ falls to
a minimum near $f^{*}$ and rises beyond it.
\item [{H4~(Correlation,~not~capability).}] Private randomness lowers
$\rho_{\text{FF}}$ and moves $\Delta(f)$ beyond $f^{*}$ from the
herding curve towards the sharing curve. It restores a positive advantage
if frontier agents randomise mainly when crowded (Corollary \ref{cor:randomisation}),
whereas a fixed rate across shares lowers $\hat{q}$ and $\hat{\epsilon}$,
shrinks the plateau, and leaves a smaller loss beyond $f^{*}$ (Remarks
\ref{rem:too-many} and \ref{rem:random-crowd}).
\end{description}

\subsection{Pilot design}\label{sec:design-factors}

\begin{table}
\caption{Design of the pilot.}\label{tab:design}

\centering{}%
\begin{tabular}{ll}
\hline 
\textbf{Factor} & \textbf{Level}\tabularnewline
\hline 
Frontier model & Gemini 3.1 Pro, low reasoning effort\tabularnewline
Secondary model & Gemini 3.1 Flash-Lite\tabularnewline
Temperature & 1.0\tabularnewline
Population size $N$ & 11\tabularnewline
Frontier count $k$ & 0, 1, 2, 3, 6, 11\tabularnewline
Memory $m$ & 3\tabularnewline
Private randomness & off, on\tabularnewline
Rounds per run & 100, with the first 50 discarded\tabularnewline
Seeds & 1 to 3 per cell\tabularnewline
\hline 
\end{tabular}
\end{table}

Table \ref{tab:design} lists the design of the pilot. The frontier
and secondary models come from one family, which reduces the differences
between the two tiers other than capability. Each combination of a
frontier count $k$ and a randomness condition is a cell, and Table
\ref{tab:pilot} gives the seeds per cell.

The endpoints $f=0$ and $f=1$ are homogeneous populations. They
provide the baseline mean payoffs and volatility for each model on
its own, and the $f=0$ population supplies $\hat{\epsilon}$.

We measure the threshold as $\hat{f}^{*}=\inf\{f:\Delta(f)\le\tau\}$
for a small tolerance $\tau$, interpolating $\Delta$ linearly between
the shares on the grid, with confidence intervals by bootstrap over
seeds. We say the advantage reverses if $\Delta(f)<-\tau$ for some
$f>\hat{f}^{*}$. The pilot uses $\tau=0.05$.

The ablation condition adds the prompt line of Listing \ref{lst:random}.
It is designed to break the symmetry between identical agents without
changing the model, and it covers every frontier count except $k=11$.

The pilot made 31,900 calls, with no request errors and no parse failures.
Mean payoffs in the two halves of the analysed window differ by no
more than sampling noise, with no consistent direction across cells.
A burn-in of 50 rounds therefore appears to suffice at this $N$.

\section{Results}\label{sec:results}

Table \ref{tab:pilot} lists the measures for every cell of the pilot.
The subsections below take the hypotheses in turn. We first estimate
$\epsilon$ and $q$, and we then compare the advantage with the benchmarks
for H1 and H2, examine the correlations and the volatility for H3,
and report the private-randomness ablation for H4.

\begin{table*}
\caption{Pilot results at $N=11$, $m=3$, Gemini 3.1 Pro against Gemini 3.1
Flash-Lite, 50 analysed rounds per seed. Rand.: private randomness
off or on.}\label{tab:pilot}

\centering{}%
\begin{tabular}{lcccccccccc}
\hline 
{\footnotesize Rand.} & {\footnotesize$k$} & {\footnotesize$f$} & {\footnotesize seeds} & {\footnotesize$\bar{w}_{\text{F}}$} & {\footnotesize$\bar{w}_{\text{S}}$} & {\footnotesize$\Delta$} & {\footnotesize$\sigma^{2}/N$} & {\footnotesize$\rho_{\text{FF}}$} & {\footnotesize$\rho_{\text{SS}}$} & {\footnotesize$\rho_{\text{FS}}$}\tabularnewline
\hline 
{\footnotesize off} & {\footnotesize 0} & {\footnotesize 0} & {\footnotesize 3} &  & {\footnotesize$-0.26$} &  & {\footnotesize$1.00$} &  & {\footnotesize$0.00$} & \tabularnewline
{\footnotesize off} & {\footnotesize 1} & {\footnotesize 0.09} & {\footnotesize 3} & {\footnotesize$-0.13$} & {\footnotesize$-0.25$} & {\footnotesize$+0.11$} & {\footnotesize$0.89$} &  & {\footnotesize$0.00$} & {\footnotesize$-0.05$}\tabularnewline
{\footnotesize off} & {\footnotesize 2} & {\footnotesize 0.18} & {\footnotesize 3} & {\footnotesize$-0.26$} & {\footnotesize$-0.24$} & {\footnotesize$-0.02$} & {\footnotesize$0.96$} & {\footnotesize$0.24$} & {\footnotesize$0.00$} & {\footnotesize$-0.01$}\tabularnewline
{\footnotesize off} & {\footnotesize 3} & {\footnotesize 0.27} & {\footnotesize 3} & {\footnotesize$-0.27$} & {\footnotesize$-0.24$} & {\footnotesize$-0.04$} & {\footnotesize$0.97$} & {\footnotesize$0.22$} & {\footnotesize$0.00$} & {\footnotesize$-0.03$}\tabularnewline
{\footnotesize off} & {\footnotesize 6} & {\footnotesize 0.55} & {\footnotesize 2} & {\footnotesize$-0.46$} & {\footnotesize$-0.13$} & {\footnotesize$-0.33$} & {\footnotesize$1.60$} & {\footnotesize$0.25$} & {\footnotesize$0.02$} & {\footnotesize$-0.02$}\tabularnewline
{\footnotesize off} & {\footnotesize 11} & {\footnotesize 1} & {\footnotesize 2} & {\footnotesize$-0.40$} &  &  & {\footnotesize$2.59$} & {\footnotesize$0.16$} &  & \tabularnewline
{\footnotesize on} & {\footnotesize 0} & {\footnotesize 0} & {\footnotesize 3} &  & {\footnotesize$-0.21$} &  & {\footnotesize$0.67$} &  & {\footnotesize$-0.03$} & \tabularnewline
{\footnotesize on} & {\footnotesize 1} & {\footnotesize 0.09} & {\footnotesize 3} & {\footnotesize$-0.27$} & {\footnotesize$-0.22$} & {\footnotesize$-0.04$} & {\footnotesize$0.83$} &  & {\footnotesize$-0.01$} & {\footnotesize$-0.03$}\tabularnewline
{\footnotesize on} & {\footnotesize 2} & {\footnotesize 0.18} & {\footnotesize 3} & {\footnotesize$-0.28$} & {\footnotesize$-0.24$} & {\footnotesize$-0.04$} & {\footnotesize$0.93$} & {\footnotesize$0.01$} & {\footnotesize$0.00$} & {\footnotesize$-0.01$}\tabularnewline
{\footnotesize on} & {\footnotesize 3} & {\footnotesize 0.27} & {\footnotesize 3} & {\footnotesize$-0.24$} & {\footnotesize$-0.26$} & {\footnotesize$+0.03$} & {\footnotesize$1.07$} & {\footnotesize$0.21$} & {\footnotesize$0.03$} & {\footnotesize$-0.04$}\tabularnewline
{\footnotesize on} & {\footnotesize 6} & {\footnotesize 0.55} & {\footnotesize 1} & {\footnotesize$-0.37$} & {\footnotesize$-0.14$} & {\footnotesize$-0.24$} & {\footnotesize$1.06$} & {\footnotesize$0.09$} & {\footnotesize$-0.04$} & {\footnotesize$-0.02$}\tabularnewline
\hline 
\end{tabular}
\end{table*}

\subsection{Estimating the exploitability and reliability}\label{sec:results-estimation}

The second test of Section \ref{sec:bench-predictions} compares the
measured threshold with the one that (\ref{eq:fstar}) predicts from
$\epsilon$ and $q$. To keep the prediction independent of $\Delta(f)$,
we estimate the two parameters from the conditional means (\ref{eq:cond-means}).

By (\ref{eq:cond-means}) and (\ref{eq:predictability}), the two
parameters follow from the predictabilities as $\epsilon=h_{\text{S}}(\mu)/2$
and $q=\tfrac{1}{2}+h_{\text{F}}(\mu)/2$. We estimate $h_{\text{S}}(\mu)$
from the $k=0$ cell, where the crowd plays alone, and $h_{\text{F}}(\mu)$
from the $k=1$ cell, where a single frontier agent faces the crowd.

The plug-in estimate of $h_{g}(\mu)$ is the absolute value of the
group's mean action over the rounds with history $\mu$. It is biased
upwards, because chance deviations of either sign count as signal.
We therefore estimate the square $h_{g}(\mu)^{2}$, which admits an
unbiased estimator, and take the square root afterwards.

Let $\bar{a}_{g}(t)$ be the mean action of group $g$ in round $t$,
and let $T_{\mu}$ be the number of analysed rounds in which history
$\mu$ occurred. Given $\mu$, the group means of two different rounds
are independent, and their product has expectation $\mathbb{E}[\bar{a}_{g}(t)\,\bar{a}_{g}(t')\mid\mu]=\mathbb{E}[a_{g}\mid\mu]^{2}=h_{g}(\mu)^{2}$.
We average this product over all pairs of different rounds, which
gives a U-statistic \cite{hoeffding1948class},
\begin{equation}
U_{g}(\mu)=\frac{1}{T_{\mu}(T_{\mu}-1)}\sum_{t\neq t'}\bar{a}_{g}(t)\,\bar{a}_{g}(t'),\label{eq:u-stat}
\end{equation}
where the sum runs over the $T_{\mu}$ rounds with history $\mu$.
The statistic is unbiased for $h_{g}(\mu)^{2}$. Because $U_{g}(\mu)$
can be negative when the lean is weak, we clip it at zero before taking
the root,
\begin{equation}
\hat{h}_{g}(\mu)=\sqrt{\max\{0,\,U_{g}(\mu)\}}.\label{eq:h-hat}
\end{equation}
A history that occurs in a single round offers no pair, and we set
$\hat{h}_{g}(\mu)=0$ for it.

Averaging $\hat{h}_{g}$ over the $T$ analysed rounds of a seed,
so that each history is weighted by its frequency, gives
\begin{equation}
\hat{\epsilon}=\frac{1}{2T}\sum_{t}\hat{h}_{\text{S}}(\mu(t)),\qquad\hat{q}=\frac{1}{2}+\frac{1}{2T}\sum_{t}\hat{h}_{\text{F}}(\mu(t)).\label{eq:eps-q-hat}
\end{equation}
We evaluate (\ref{eq:h-hat}) and (\ref{eq:eps-q-hat}) within each
seed and average the results over the seeds of a cell. In the pilot,
the $k=0$ cell gives $\hat{\epsilon}=0.039$ and the $k=1$ cell
gives $\hat{q}=0.737$. Equation (\ref{eq:fstar}) then predicts $f^{*}=0.14$
with nothing fitted.

The estimators converge to the true values as the number of rounds
grows, but the pilot is small, and its size limits the estimates in
three ways. First, the predicted $f^{*}$ has a 95\% range of 0.07
to 0.25 at this size. The range comes from simulating the herding
benchmark at these estimates, with three seeds of 50 analysed rounds
per cell. Second, taking the root biases the estimates downwards at
this size, with a true $q$ of 0.737 giving $\hat{q}=0.69$ on average
in the same simulation. Third, the crowd's lean is only just distinguishable
from none. In the same simulation a crowd with $\epsilon=0$ gives
$\hat{\epsilon}=0.022$ on average, and reaches 0.039 in about 2\%
of runs.

\begin{figure}
\begin{centering}
\includegraphics[width=1\columnwidth]{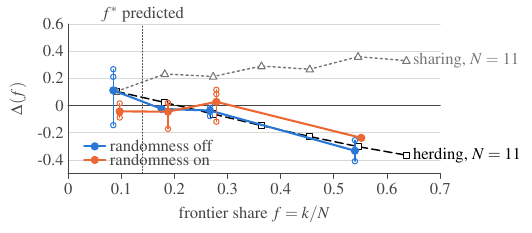}
\par\end{centering}
\caption{Pilot frontier advantage $\Delta(f)$ at $N=11$, with private randomness
off (blue) and on (orange). Small open circles are single seeds, large
filled circles are the means over the seeds of a cell, and the bars
are the seed range. The solid lines join the means. The squares are
the exact herding values at $N=11$, at $\hat{\epsilon}=0.039$ and
$\hat{q}=0.737$ from (\ref{eq:eps-q-hat}), and the triangles are
the finite-$N$ sharing values of Section \ref{sec:bench-predictions}
at the same parameters. The vertical dotted line marks the predicted
$f^{*}$.}\label{fig:results-delta}
\end{figure}

\subsection{Frontier advantage versus frontier share}\label{sec:results-dilution}

Fig. \ref{fig:results-delta} shows $\Delta(f)$ at the four mixed
shares, with the two benchmarks evaluated at the estimated parameters,
and Table \ref{tab:results-delta} lists the values. With randomness
off, $\Delta$ is positive only at $k=1$, where the mean is $+0.11$.
The advantage is relative, because the frontier agent itself still
loses 0.13 per round. At $k=2$ and $k=3$ the advantage has vanished,
with $\Delta=-0.02$ and $-0.04$, both within the tolerance $\tau=0.05$
of zero. Interpolating between $k=1$ and $k=2$ gives a measured
$\hat{f}^{*}=0.13$. The 95\% bootstrap interval over seeds is 0.09
to 0.16, and its lower end is the smallest share on the grid, $f=1/11$,
reached when a resample puts the $k=1$ cell below $\tau$. The pilot
is therefore consistent with H1.

The point estimate is also close to the predicted 0.14, which is the
threshold part of H3. Given the range of the prediction, this agreement
is a consistency check. The two are also defined differently. The
prediction is the root of the drift, whereas the measurement is the
share at which $\Delta$ falls to $\tau$ at $N=11$. Applying the
same interpolation to the exact herding values gives 0.15, and the
measured 0.13 is close to that as well.

At $k=6$ both seeds are negative, $-0.25$ and $-0.41$, with a mean
of $-0.33$. The advantage has therefore reversed by a share of 0.55.
The grid has no share between 0.27 and 0.55, and the reversal may
begin anywhere in that gap. More importantly, the two benchmarks predict
opposite signs at $k=6$. Herding gives $-0.30$ and sharing gives
$+0.36$, and the measured $-0.33$ favours herding, as H2 predicts.

The exact herding values are also close to the measurements at $k=1$,
2 and 3, with a largest gap of 0.05. The agreement does not come from
errors that cancel in the difference. The same enumeration gives the
two payoffs separately, and they lie within 0.04 of the measured $\bar{w}_{\text{F}}$
and $\bar{w}_{\text{S}}$ at every mixed share. The sharing values
are computed at $N=11$ as well (Section \ref{sec:bench-predictions}).
With a single frontier agent sharing equals herding, and the $k=1$
cell therefore cannot separate them. At $k=2$ and $k=3$ sharing
gives $+0.23$ and $+0.21$, whereas the measured values are $-0.02$
and $-0.04$ and no seed is above 0.01. The measurements therefore
lie on the herding side at every share with more than one frontier
agent.

The reversal does not by itself show that the frontier model reads
the crowd. The unaligned rule of Remark \ref{rem:unaligned} gives
$-0.28$ at $k=6$, close to the herding value of $-0.30$, because
a bloc of six is the majority whichever way it leans. The two differ
most at $k=1$, where herding gives $+0.10$ and the unaligned rule
$+0.01$. The measured $+0.11$ lies on the herding side, but its
standard error over seeds is 0.13.

\begin{table}
\caption{Measured frontier advantage against the benchmarks at $N=11$ with
randomness off. Herding is the exact value at $N=11$, found by enumeration
(Remark \ref{rem:finite-n}). Unaligned is the same enumeration for
the rule of Remark \ref{rem:unaligned}. Sharing is the finite-$N$
value of Section \ref{sec:bench-predictions}. All are at $\hat{\epsilon}=0.039$
and $\hat{q}=0.737$.}\label{tab:results-delta}

\centering{}%
\begin{tabular}{cccccc}
\hline 
$k$ & $f$ & \textbf{Measured} & \textbf{Herding} & \textbf{Unaligned} & \textbf{Sharing}\tabularnewline
\hline 
$1$ & $0.09$ & $+0.11$ & $+0.10$ & $+0.01$ & $+0.10$\tabularnewline
$2$ & $0.18$ & $-0.02$ & $+0.02$ & $-0.05$ & $+0.23$\tabularnewline
$3$ & $0.27$ & $-0.04$ & $-0.06$ & $-0.11$ & $+0.21$\tabularnewline
$6$ & $0.55$ & $-0.33$ & $-0.30$ & $-0.28$ & $+0.36$\tabularnewline
\hline 
\end{tabular}
\end{table}

\subsection{Mechanism: correlation and volatility}\label{sec:results-mechanism}

\begin{figure}
\begin{centering}
\includegraphics[width=1\columnwidth]{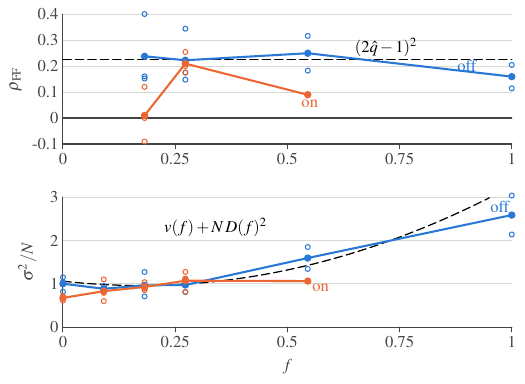}
\par\end{centering}
\caption{Pilot mechanism measures at $N=11$. Top: within-frontier correlation
$\rho_{\text{FF}}$ for $k\ge2$, with the benchmark value $(2\hat{q}-1)^{2}$.
Bottom: volatility $\sigma^{2}/N$, with the herding prediction $v(f)+N\,D(f)^{2}$
at $\hat{\epsilon}$ and $\hat{q}$. Blue: randomness off. Orange:
randomness on. Small circles are seeds.}\label{fig:results-mech}
\end{figure}

The correlations $\rho_{\text{FF}}$, $\rho_{\text{SS}}$ and $\rho_{\text{FS}}$
are the mean pairwise action correlations within and between the two
groups, and the volatility is $\sigma^{2}/N$ as defined in Section
\ref{sec:bench-predictions}. Fig. \ref{fig:results-mech} shows $\rho_{\text{FF}}$
and $\sigma^{2}/N$, and Table \ref{tab:pilot} lists the values.
The correlation $\rho_{\text{FF}}$ is 0.24, 0.22 and 0.25 at $k=2$,
3 and 6, while $\rho_{\text{SS}}$ stays between 0.00 and 0.02. The
benchmark implies $\rho_{\text{FF}}=(2\hat{q}-1)^{2}=0.22$ and $\rho_{\text{SS}}=4\hat{\epsilon}^{2}=0.006$,
and the measured values are close to both.

The cross-correlation $\rho_{\text{FS}}$ is the signed check, because
the unaligned rule gives zero. It is slightly negative, between $-0.01$
and $-0.05$, in line with the benchmark's $-0.04$, and it is negative
in 9 of the 11 seeds without randomness. A more direct check gives
the same weak answer. Frontier choices oppose the crowd's lean in
54\% of cases, with a bootstrap 95\% interval of 51\% to 58\%, between
the 60\% that herding would give on this measure and the 50\% of the
unaligned rule.

The volatility has the shape that herding predicts. It dips from 1.00
at $k=0$ to 0.89 at $k=1$ and rises to 1.60 at $k=6$, where herding
gives 1.42 and sharing gives 0.50. This is not evidence beyond the
correlations, however, because identity (\ref{eq:vol-identity}) gives
1.61 at $k=6$ from the three measured correlations. An all-frontier
population loses 0.40 per agent per round, against 0.26 for an all-secondary
population.

At $k=11$ both measures fall short of the benchmark. The correlation
is 0.16 against 0.22, and the volatility is 2.59 against 3.25. Frontier
agents therefore herd less in an all-frontier population. Taken together,
the evidence for H3 is suggestive rather than conclusive.

\subsection{Private-randomness ablation}\label{sec:results-randomness}

Giving every agent a private random number decorrelated the frontier
agents at two of the three shares, as Table \ref{tab:pilot} and the
orange line of Fig. \ref{fig:results-mech} show. The correlation
$\rho_{\text{FF}}$ fell from 0.24 to 0.01 at $k=2$ and from 0.25
to 0.09 at $k=6$, but stayed at 0.21 at $k=3$. Volatility at $k=6$
fell from 1.60 to 1.06.

Randomness did not restore the advantage at $k\ge2$, and it also
removed the advantage at $k=1$, where $\Delta$ fell from $+0.11$
to $-0.04$. Both terms of the plateau fell, with $\hat{q}$ down
from 0.737 to 0.652 and $\hat{\epsilon}$ down from 0.039 to 0.030,
and (\ref{eq:fstar}) at these estimates raises the predicted threshold
from 0.14 to 0.17. The measured threshold cannot be compared with
this prediction, because the advantage is already within the tolerance
at the smallest share on the grid, $f=1/11$. At $k=6$ the single
seed gives $-0.24$, a smaller loss than the $-0.33$ without randomness,
as H4 predicts, but one seed cannot separate it from the two seeds
there.

This pattern is what Remark \ref{rem:too-many} predicts for a fixed
rate, used at every share rather than only when crowded. The logs
do not support reading the fall as coin flips alone, however. Frontier
choices match the rule that plays A when the number is below one half
in 58\% of calls, and secondary choices in 52\%, against 50\% by chance.
A fall in $\hat{q}$ from 0.737 to 0.652 would need about a third
of frontier choices to follow the number, whereas a 58\% match implies
about a sixth. The fall in $\hat{q}$ may therefore come partly from
a change in the frontier rule.

The secondary crowd changed as well, because the ablation gives every
agent a number. On its own it improved from $-0.26$ to $-0.21$ per
round, with $\sigma^{2}/N=0.67$, below the value of 1 for independent
coin flips. The secondary agents did not reach this by using the number
as a coin. They mention it in 88\% of their rationales, but following
the rule above would give $\sigma^{2}/N=0.98$. Their actions are
therefore not independent given the history, as the benchmarks assume.

The pilot supports H4 only in part. Private randomness lowered $\rho_{\text{FF}}$
at two of the three shares, but it did not restore a positive advantage,
and the smaller loss at $k=6$ rests on a single seed. The ablation
also changes both groups at once, and a variant that gives the number
only to frontier agents would isolate the effect on the frontier agents.

\section{Discussion}\label{sec:discussion}

Two points stand out from the pilot. First, the data lie on the herding
side (Fig. \ref{fig:results-delta} and Table \ref{tab:results-delta}).
The frontier agents follow the same rule and act on it together, as
if none of them expected another agent to reason in the same way.
The frontier agent is predictable. Its lean given the history is 0.24
above chance, since $\hat{q}=0.74$, while the crowd leans by only
$\hat{\epsilon}=0.04$. Whether that lean reads the crowd, in the
sense of opposing the side the crowd tilts to, is less certain. The
cross-correlation and the edge of the lone frontier agent both point
that way, but an unaligned rule fits the other measurements nearly
as well (Remark \ref{rem:unaligned}). Beyond the threshold the distinction
matters little, because a bloc that large is the majority whichever
way it leans. Its loss then grows with its reliability, as Proposition
\ref{prop:herding} shows.

Second, private randomness lowered the correlation at two of the three
shares (Fig. \ref{fig:results-mech}) but did not restore the advantage,
and it removed the edge of the lone frontier agent. One reading is
that agents randomise when they are rare as well as when they are
crowded, since Corollary \ref{cor:randomisation} and Remark \ref{rem:too-many}
say the useful rate depends on $f$. The logs qualify this reading,
because few choices follow the number and the frontier rule itself
may have changed (Section \ref{sec:results-randomness}). A randomisation
policy that an agent conditions on its own recent losses, or model
diversity, may do better and remains to be tested.

Four caveats apply. First, the results come from one model pair and
one prompt at $N=11$ and $m=3$, with one to three seeds per cell,
and the bootstrap interval for $\hat{f}^{*}$ rests on two or three
seeds per share. Second, the threshold estimate depends on the tolerance.
With $\tau=0$ the interpolated threshold is 0.17 in place of 0.13,
against 0.21 in place of 0.15 for the exact herding values. Third,
the benchmarks hold $\epsilon$ and $q$ constant in $f$ and fix
a single lean per history, and they therefore describe play before
agents adapt. Fourth, the frontier and secondary tiers differ in more
than capability, which within-family pairs reduce but do not remove.

\section{Conclusion}\label{sec:conclusion}

This paper asked what a frontier model is worth in the Minority Game
once many competitors run it too. We framed the question as the frontier
advantage $\Delta(f)$ against the frontier share $f$, and we derived
two benchmarks that share the closed-form threshold of (\ref{eq:fstar})
but disagree beyond it. Herding predicts a reversal, and sharing predicts
a positive edge that decays as $1/f$.

A pilot with 11 agents, Gemini 3.1 Pro against Gemini 3.1 Flash-Lite,
lies on the herding side. A single frontier agent has a small and
noisy edge, the edge is gone with two or three, and six frontier agents
lose to the secondary agents they outnumber. The reversal needs only
a shared rule, however, and private randomness did not restore the
advantage. These results are consistent with the hypotheses but do
not confirm them, because they rest on one model pair, one prompt
and one to three seeds per cell.

If the pattern holds at larger $N$ and for other model pairs, it
carries a clear lesson. A frontier model pays off while few competitors
run it. Once adoption passes $f^{*}$, however, even perfect sharing
gains nothing more from capability (Remark \ref{rem:ceiling}), and
herding turns the edge into a loss (Proposition \ref{prop:herding}).
That loss comes from sharing one model rather than having a stronger
one, which a middle-tier bloc at the same share would test. Likewise,
private randomness avoids the loss only when its rate rises with adoption
(Corollary \ref{cor:randomisation} and Remark \ref{rem:too-many}).
Where autonomous agents compete for scarce capacity, advantage therefore
shifts from deploying the strongest model to not thinking alike.

\bibliographystyle{ieeetr}
\bibliography{bibliography}

\begin{thebibliography}{10}

\bibitem{arthur1994}
W.~B. Arthur, ``Inductive reasoning and bounded rationality,'' {\em American
  Economic Review}, vol.~84, no.~2, pp.~406--411, 1994.

\bibitem{challet1997}
D.~Challet and Y.-C. Zhang, ``Emergence of cooperation and organization in an
  evolutionary game,'' {\em Physica A}, vol.~246, pp.~407--418, 1997.

\bibitem{johnson1999}
N.~F. Johnson, M.~Hart, and P.~M. Hui, ``Crowd effects and volatility in
  markets with competing agents,'' {\em Physica A}, vol.~269, no.~1, pp.~1--8,
  1999.

\bibitem{challet2004}
D.~Challet, M.~Marsili, and G.~Ottino, ``Shedding light on {El Farol},'' {\em
  Physica A}, vol.~332, pp.~469--482, 2004.

\bibitem{challet2000}
D.~Challet, M.~Marsili, and Y.-C. Zhang, ``Modeling market mechanism with
  minority game,'' {\em Physica A}, vol.~276, pp.~284--315, 2000.

\bibitem{johnson1999mixed}
N.~F. Johnson, P.~M. Hui, D.~Zheng, and M.~Hart, ``Enhanced winnings in a
  mixed-ability population playing a minority game,'' {\em Journal of Physics
  A: Mathematical and General}, vol.~32, no.~38, pp.~L427--L431, 1999.

\bibitem{grossman1980}
S.~J. Grossman and J.~E. Stiglitz, ``On the impossibility of informationally
  efficient markets,'' {\em American Economic Review}, vol.~70, no.~3,
  pp.~393--408, 1980.

\bibitem{kleinberg2021}
J.~Kleinberg and M.~Raghavan, ``Algorithmic monoculture and social welfare,''
  {\em Proceedings of the National Academy of Sciences}, vol.~118, no.~22,
  p.~e2018340118, 2021.

\bibitem{akata2025}
E.~Akata, L.~Schulz, J.~Coda-Forno, S.~J. Oh, M.~Bethge, and E.~Schulz,
  ``Playing repeated games with large language models,'' {\em Nature Human
  Behaviour}, vol.~9, pp.~1380--1390, 2025.

\bibitem{huang2025}
J.-t. Huang, E.~J. Li, M.~H. Lam, T.~Liang, W.~Wang, Y.~Yuan, W.~Jiao, X.~Wang,
  Z.~Tu, and M.~R. Lyu, ``Competing large language models in multi-agent gaming
  environments,'' in {\em International Conference on Learning Representations
  (ICLR)}, 2025.

\bibitem{mori2026}
D.~M. Mori and N.~F. Johnson, ``Three {AI}-agents walk into a bar \ldots {`Lord
  of the Flies'} tribalism emerges among smart {AI}-agents,'' {\em arXiv
  preprint arXiv:2602.23093}, 2026.

\bibitem{takata2025}
R.~Takata, A.~Masumori, and T.~Ikegami, ``Emergent social dynamics of {LLM}
  agents in the {El Farol} bar problem,'' {\em arXiv preprint
  arXiv:2509.04537}, 2025.

\bibitem{ballestero2026}
G.~Ballestero, H.~Hosseini, S.~Khanna, and R.~I. Shorrer, ``Strategic
  algorithmic monoculture: {Experimental} evidence from coordination games,''
  {\em arXiv preprint arXiv:2604.09502}, 2026.

\bibitem{sinishaw2026}
D.~Sinishaw, Q.~Zhu, E.~Meriaux, and G.~Dudek, ``Do {LLMs} beat {Nash}?
  {Testing} decentralized coordination in self-play multi-agent games,'' {\em
  arXiv preprint arXiv:2608.12547}, 2026.

\bibitem{ross2026}
J.~Ross, E.~So, Z.~De~Simone, C.~Pozniak, and A.~W. Lo, ``Why better models can
  create riskier systems: {Evidence} from {LLM} agents in financial markets,''
  {\em arXiv preprint arXiv:2609.04373}, 2026.

\bibitem{challet2000stat}
D.~Challet, M.~Marsili, and R.~Zecchina, ``Statistical mechanics of systems
  with heterogeneous agents: Minority games,'' {\em Physical Review Letters},
  vol.~84, pp.~1824--1827, 2000.

\bibitem{hoeffding1948class}
W.~Hoeffding, ``A class of statistics with asymptotically normal
  distribution,'' {\em The Annals of Mathematical Statistics}, vol.~19, no.~3,
  pp.~293--325, 1948.

\end{thebibliography}

\end{document}